\documentclass[conference]{IEEEtran}
\IEEEoverridecommandlockouts
\usepackage{cite}
\usepackage{amsmath,amssymb,amsfonts}
\usepackage{algorithmic}
\usepackage{graphicx}
\usepackage{textcomp,adjustbox}
\usepackage{xcolor}
\usepackage{relsize}
\usepackage{subcaption}
\usepackage{epsfig}
\usepackage{multirow}
\usepackage{comment}
\usepackage[ruled]{algorithm2e}
\usepackage{amsmath}

\usepackage{bbm}

\def\BibTeX{{\rm B\kern-.05em{\sc i\kern-.025em b}\kern-.08em
    T\kern-.1667em\lower.7ex\hbox{E}\kern-.125emX}}

\begin{document}
\title{Leveraging Semi-Supervised Learning for Fairness using Neural Networks}

\author{\IEEEauthorblockN{Vahid Noroozi}
\IEEEauthorblockA{\textit{Department of Computer Science} \\
\textit{University of Illinois at Chicago}\\
Chicago, IL, USA \\ 
vnoroo2@uic.edu}

\\

\IEEEauthorblockN{Nooshin Mojab}
\IEEEauthorblockA{\textit{Department of Computer Science} \\
\textit{University of Illinois at Chicago}\\
Chicago, IL, USA \\
nmojab2@uic.edu}

\and

\IEEEauthorblockN{Sara Bahaadini}
\IEEEauthorblockA{\textit{Department of Computer Science} \\
\textit{Northwestern University}\\
Evanston, IL, USA \\
sara.bahaadini@u.northwestern.edu}

\\

\IEEEauthorblockN{Philip S. Yu}
\IEEEauthorblockA{\textit{Department of Computer Science} \\
 \textit{University of Illinois at Chicago}\\
Chicago, IL, USA \\
psyu@uic.edu}

\and

\IEEEauthorblockN{Samira Sheikhi}
\IEEEauthorblockA{\textit{Zillow Group} \\
Seattle, WA, USA \\
samiras@zillowgroup.com \\
}

\\
\\

}

\maketitle

\begin{abstract}
There has been a growing concern about the fairness of decision-making systems based on machine learning. The shortage of labeled data has been always a challenging problem facing machine learning based systems. In such scenarios, semi-supervised learning has shown to be an effective way of exploiting unlabeled data to improve upon the performance of model. Notably, unlabeled data do not contain label information which itself can be a significant source of bias in training machine learning systems. This inspired us to tackle the challenge of fairness by formulating the problem in a semi-supervised framework. In this paper, we propose a semi-supervised algorithm using neural networks benefiting from unlabeled data to not just improve the performance but also improve the fairness of the decision-making process. The proposed model, called SSFair, exploits the information in the unlabeled data to mitigate the bias in the training data.

\end{abstract}


\section{Introduction}

The rapid increase in automation of decision-making systems using machine learning approaches has raised significant concerns about the fairness of such models. Different studies have shown that these machine learning models which are designed to help the process of decision-making are not immune to social biases  \cite{barocas2016big,chouldechova2017fair}. There is a significant shift towards employing machine learning techniques in many sensitive real-world applications such as credit approval, loan applications, criminal risk assessment, university admissions, and online advertisement. With this new trend, it becomes crucial to consider other aspects and metrics for assessing a model beyond their accuracy. Among those aspects, fairness has gathered close attention in the community as we hope for building a socially responsible and inclusive system. Recently many machine learning algorithms have been proposed to address this problem and make the predictions of the learning algorithms fairer \cite{hardt2016equality,louizos2016variational,agarwal2018reductions,zafar2017fairness2}.

The naive approach for addressing the fairness problem in machine learning could be to remove or ignore the protected attributes such as sex, gender and age. However, this approach is not practical in many real-world applications \cite{lum2016statistical} mainly because of the two following reasons: 1) there can exist some proxy features or correlation between other features and the sensitive attributes which may reveal them, and 2) there already exists some degree of bias in the labels of the training data. On the other hand, in many applications, unlabeled data is abundant, and if appropriately leveraged, they also hold less bias compared to labeled data since models are not strongly affected by the labels of the labeled samples. In the same way, other paradigms like unsupervised learning or semi-supervised learning could be to lower degree sensitive to these biases in the data. Additionally, the lack of adequate labeled data poses a major challenge to many machine learning based applications and in some applications, creating a labeled dataset for training such models is expensive and time-consuming. Therefore, leveraging unlabelled data could be a potential solution to the lack of labeled data as well as fairness problems.

Semi-supervised learning approaches have shown promising results in tackling the aforementioned challenges by exploiting the unlabeled data to improve the performance of a classifier in terms of the accuracy \cite{oliver2018realistic}. The unlabeled data do not carry label information which can be a significant source of bias in training machine learning systems. The success of semi-supervised approaches in the improvement of model's performance through exploiting the unlabeled data, inspired us to study the effect of unlabeled data on the process of learning a fair classifier. In this paper, we propose a semi-supervised classification algorithm based on neural networks to tackle the fairness in machine learning. To the best of our knowledge, we are the first to propose and study the effect of semi-supervised learning on the fairness of a classifier using neural networks. Our proposed model, called SSFair, utilizes Pseudo-Labeling \cite{lee2013pseudo} approach to exploit unlabeled data to increase the accuracy and fairness of a classifier. Pseudo-labeling is one of the most common techniques for handling semi-supervised learning.    

The proposed model is built with neural networks and can support any fairness measurement which can be defined or approximated as a differentiable function. Different criteria exist to measure fairness in machine learning. We have incorporated three of the most common measurements demographic parity, equalized opportunity, and equalized odds \cite{hardt2016equality} into SSFair. We have evaluated SSFair on different measurements of fairness in semi-supervised settings and showed the effectiveness of the proposed algorithm to exploit the unlabeled data. We show experimentally that SSFair can benefit from unlabeled data to not just improve the accuracy but also improve the fairness of the classifier.

\section{Related Works}
There are three main approaches proposed to tackle the fairness problem in machine learning, 1) pre-processing, 2) in-processing, and 3) post-processing approach. 

In pre-processing approach, the goal is to learn a new representation of the data which is uncorrelated with the protected attributes \cite{gordaliza2019obtaining,calmon2017optimized,louizos2016variational,adler2018auditing}. This new representation can be used for any downstream task such as classification or ranking and any machine learning technique of choice. The main advantage of pre-processing approach is that it eliminates the need for making changes to the machine learning algorithms and therefore is very straightforward to use.

The second approach, in-processing, consists of the techniques that incorporate the fairness constraints into the training process. Most of the works on fairness in machine learning belong to this category \cite{agarwal2018reductions, zafar2017fairness,kamiran2012decision}. The in-processing algorithms usually address the problem by adding the fairness criterion to the learning algorithm's main objective function as a regularizer. This category is more flexible to optimize different fairness constraints, and the solutions using this approach are considered the most robust ones. Moreover, these category of approaches have shown promising results in terms of both accuracy and fairness.  

The third approach is post-processing which aims to make changes on the output of the classifiers in order to satisfy the fairness constraint. One simple form of it is to find a threshold specific for each protected group and use it to control the fairness objective. Although this approach does not need any changes in the classifier, it is not very flexible in optimizing the trade-off between fairness and accuracy.

Our proposed model formulated as a semi-supervised learning based on neural network, falls under the second category, in-processing approaches. It aims at optimizing the fairness constraint during training the classifier. To the best of our knowledge, SSFair is the first semi-supervised algorithm based on neural networks introduced for tackling the fairness problem.

There are a few works that employ neural networks to optimize the trade-off between fairness and accuracy. Most of these approaches employ adversarial optimization inspired by Generative Adversarial Networks (GAN) \cite{goodfellow2014generative} to train a model for producing a fair representation or an output which is indistinguishable among all of the protected groups \cite{louppe2017learning,celis2019improved,wadsworth2018achieving,zhang2018mitigating,madras2018learning}. However, these methods are not capable of optimizing an arbitrary fairness constraint, at least not explicitly.
Alternatively, in \cite{manisha2018neural} fairness problem is addressed by incorporating the fairness constraints explicitly into the optimization of the neural network during the training. The authors have added several fairness constraints into the loss function of the neural network as a regularization term. This algorithm only handles fully supervised learning setting and thus can not benefit from unlabeled data.

\section{Fairness Measurements}
\label{fairconstraints}
Defining the concept of fairness for a machine learning algorithm is not trivial, and a variety of definitions exist to measure and quantify fairness. \cite{hardt2016equality,kim2018fairness}. Such definitions are categorized into two main groups of individual fairness \cite{dwork2012fairness} and group fairness \cite{hardt2016equality}. 

The term of individual fairness is first introduced in \cite{dwork2012fairness} to refer to a fairness constraint which is focused on treating similar individuals as similar as possible. The fairness measurement or metrics defined in this category are based on the expectation that similar individuals should get treated similarly and the output of the machine learning algorithm should be close for similar inputs \cite{kim2018fairness,zemel2013learning}. The main drawback of such constraints is the difficulty of defining their similarly metric function. An appropriate similarity function should be capable of ignoring the proxy features which may reveal individual's sensitive information. For this reason, individual fairness cannot be applied widely in real-world problems.

The second group, called group fairness or statistical fairness, is most commonly used in the literature. They divide the individuals or samples into sets of unprotected and protected (or privileged and unprivileged) based on sensitive attributes like race, gender, or age. Then they try to make some statistical measures (e.g. classification error, true positive rate, or false positive rate) of the performance of the classifier or any other machine learning algorithm equal for both the protected and unprotected groups. The three most common definitions in this category are demographic parity, equalized opportunity, and equalized odds. Our SSFair approach can optimize for all of these three fairness objectives. These measurements are defined in Section \ref{proposedmodel} in detail.

There is no consensus on the best definition of fairness, and it is very task-dependent to decide which one to use. In some cases, there exists a trade-off between some of these fairness constraints. It is shown that some of these fairness constraints cannot get satisfied at the same time except in some degenerate or highly constrained special cases \cite{kleinberg2017inherent,pleiss2017fairness}.
\section{Proposed Model}
\label{proposedmodel}
In semi-supervised settings, training data consists of a collection of labeled and unlabeled samples. Assume ${\mathcal{D}} = \big\{(X_i,a_i,y_i) \big\}_{i=1}^{N}$ is the training set consisting of $N$ samples. For each sample $i$, $X_i$ denotes the feature set, $y_i \in \{ 0, 1, u\}$ denotes the label, and $a_i \in \{p, n\}$ is the protected attribute which shows whether that sample belongs to the protected set ($p$) or not ($n$). Assume the valid values for labels are $0$ for non-advantaged outcome, $1$ for the advantaged outcome, or $u$ for the unknown labels.

Our goal is to learn a binary classifier function $f(X;\Theta):\cal{X}\xrightarrow{}\cal{Y}$ parameterized by $\Theta$ to optimize two main objectives, the classification accuracy, and fairness. We would model the function $f(.)$ by a neural network. To achieve this goal we define the loss function of the model as:

\begin{equation}
\mathcal{J}(\mathcal{D};{\Theta}) =  
\alpha \mathcal{J_C}(\mathcal{D};{\Theta}) + 
(1 - \alpha) \mathcal{J_F}(\mathcal{D};{\Theta}) + 
\beta \left\| {{\Theta}} \right\|_2
\label{eq:eq1}
\end{equation}
\noindent
where $\mathcal{J_C}(\mathcal{D};{\Theta})$ indicates the classification loss, and $\mathcal{J_F}(D;{\Theta})$ is the fairness loss which imposes fairness on the output of the model. Parameter $\alpha$ controls the trade-off between fairness and accuracy losses. Parameter $\beta$ controls the regularization term $\|\Theta\|$ which is imposed on all of the networks' weights. Regularization is very important to prevent overfitting specially since limited labeled samples are available.
\subsection{Classification Loss}
The first part of the loss function, the classification accuracy loss $\mathcal{J_C}(\mathcal{D};{\Theta})$, is defined over the training samples as:
\begin{equation}
\mathcal{J_C}(\mathcal{D};{\Theta})=\sum\nolimits_{1\le i\le N} j_c(X_i;\Theta)
\label{eq:eq2}
\end{equation}

\noindent
where $j_c(X_i)$ indicates the classification loss for sample $X_i$ and is defined as the cross-entropy between the output of the learned function and the target label:
\begin{equation}
j_c(X_i;\Theta) = \mathbbm{1}{\{v_i=T\}} (-q_i \log {\hat{y}_i} - (1 - q_i) \log (1-\hat{y}_i))
\label{eq:eq30}
\end{equation}

\noindent
where $\hat{y}_i=f(X_i;\Theta), 0<\hat{y}_i<1$ indicates the output of the learned function for sample $X_i$ and $q_i$ is $X_i$'s corresponding target label. Target label $q_i$ is defined as the ground truth label $y_i$ if $X_i$ is labeled, while it is defined as $q_i=\mathbbm{1}{\{\hat{y}\ge 0.5\}}$ for unlabeled samples. $v_i \in \{T,F\}$ indicates whether sample $X_i$ should be considered in the learning process or not and will be defined below. $\mathbbm{1}$ is an indicator function which zero-outs the samples whose $v_i$ is not $T$.

\vspace{1mm}
We follow the Pseudo-Label approach \cite{lee2013pseudo} to handle the unlabeled samples. For all labeled samples, $v_i$ is set to $T$. For unlabeled samples, only the ones with high confidence output should get their $v_i$ set to $T$ and remain in the learning process. With a binary classifier, the output value $\hat{y}$ can be utilized to obtain the confidence of the prediction for sample $X_i$. Therefore $v_i$ is defined as:

\vspace{-5mm}
\begin{equation}
v_i = \left\{ \begin{array}{l}
T{\rm{  \quad if \quad}} y_i = 0\;or\;1 \\
T{\rm{  \quad if \quad}} y_i = u \; and \; (\hat{y}_i < (1-\lambda)\; or\; \hat{y}_i > \lambda)  \\
F{\rm{   \quad if \quad}} otherwise 
\end{array} \right.
\label{eq:eq40}
\end{equation}
\noindent
where $0\leq\lambda\leq1$ defines a threshold which controls the degree of confidence which is needed to consider a predicted label in the learning process.
\subsection{Fairness Loss}
The second term in the loss imposes fairness on the learned function. As discussed in Section \ref{fairconstraints}, there is a variety of definitions for fairness and there is no consensus on which one is the best.
Our approach is quite flexible in that it can work with any fairness objective as far as it is a differentiable function. This capacity to handle and optimize different definitions is considered a huge advantage for a fairness algorithm since it enables adapting the appropriate fairness definition based on the application. In this paper, the following three most common objectives in group fairness are studied with the proposed model.

\vspace{1.5mm}
\subsubsection{Demographic Parity}
 Demographic parity, also referred to as Statistical Parity, is one of the most common criteria for fairness \cite{barocas2016big}. It measures the difference between the probabilities of predicting advantaged output for the protected and unprotected groups and requires the decision of a classifier to be independent of the protected attribute $a$. Its corresponding loss function denoted by $\mathcal{J_F^{DP}}(\mathcal{D};{\Theta})$ is defined as:
{\small
\begin{flalign}
\mathcal{J_F^{DP}}(D;\Theta ) = \left| \mathbb{E}[f(X;\Theta|a=p)] - \mathbb{E}[f(X;\Theta|a=n)] \right|
\label{eq:eq5}
\end{flalign}}
\noindent
with:
\begin{equation}
\mathbb{E}[f(X;\Theta|a=z)] = \frac{{{\sum}_{{X_i} \in {D_{a = z}}}{f({X_i};\Theta )} }}{{\left| {{D_{a = z}}} \right|}}
\end{equation}
\noindent
where $D_{a=z}$ defines the subset of $D$ where their protected attribute $a=z$.

Demographic parity is backed up by the "four-fifth rule" which recommends that the selection rate for the protected group should not be less than $80\%$ of the unprotected group unless there exists some business necessity \cite{bobko2004four}. A selection rate of less than $80\%$ can have an adverse impact on the unprotected group. 

\vspace{1mm}
\subsubsection{Equalized Opportunity}
This measurement is focused on fairness for the advantaged outcome. It measures the difference between the probabilities of predicting advantaged output for the protected and unprotected groups with advantaged ground truth. Its corresponding loss function denoted by $\mathcal{J_F^{ODD}}(\mathcal{D};{\Theta}, k)$ is defined as the following when $k=1$:

\vspace{-2mm}
\begin{multline}
\mathcal{J_F^{OPP}}(D;\Theta ,k) = | \mathbb{E}[f(X;\Theta|a=p, y=k)] - \\ \mathbb{E}[f(X;\Theta|a=n, y=k)]|
\label{eq:eq7}
\end{multline}
\noindent
with:
\begin{equation}
\mathbb{E}[f(X;\Theta|a=z, y=k)] = \frac{{\mathlarger{\sum}\nolimits_{{X_i} \in {D_{a = z}} \cap {D_{y = k}}}^{} {f({X_i};\Theta )} }}{{\left| {{D_{a = z}} \cap {D_{y = k}}} \right|}}
\end{equation}
\noindent 
where $D_{y=k}$ defines the subset of $D$ with label attribute $y=k$.

\subsubsection{Equalized Odds}
This constraint requires the outcome and the protected attribute to be independent conditional on the label. It is defined as the following:
\begin{equation}
\mathcal{J_F^{ODD}}(D;\Theta ) = \mathcal{J_F^{OPP}}(D;\Theta ,k = 0) + \mathcal{J_F^{OPP}}(D;\Theta ,k = 1)
\label{eq:eq9}
\end{equation}

It is a more strict criterion than equalized opportunity as it requires for both $y=1$ and $y=0$. It enforces the accuracy to be equally high for all of the outcomes while equalized opportunity focuses on the advantaged outcome.

\subsection{Model and Training}
The classifier function $f(.)$ is modelled by a multi-layer perception (MLP) neural network. The whole model is trained using backpropagation with respect to the loss function in Equation \eqref{eq:eq1}. Given a set of $N$ samples, we optimize the model using Adam \cite{kingma2014adam} optimization technique over shuffled mini-batches from the data. 

\begin{figure}[ht]
    \centering
    \includegraphics[width=0.33\textwidth]{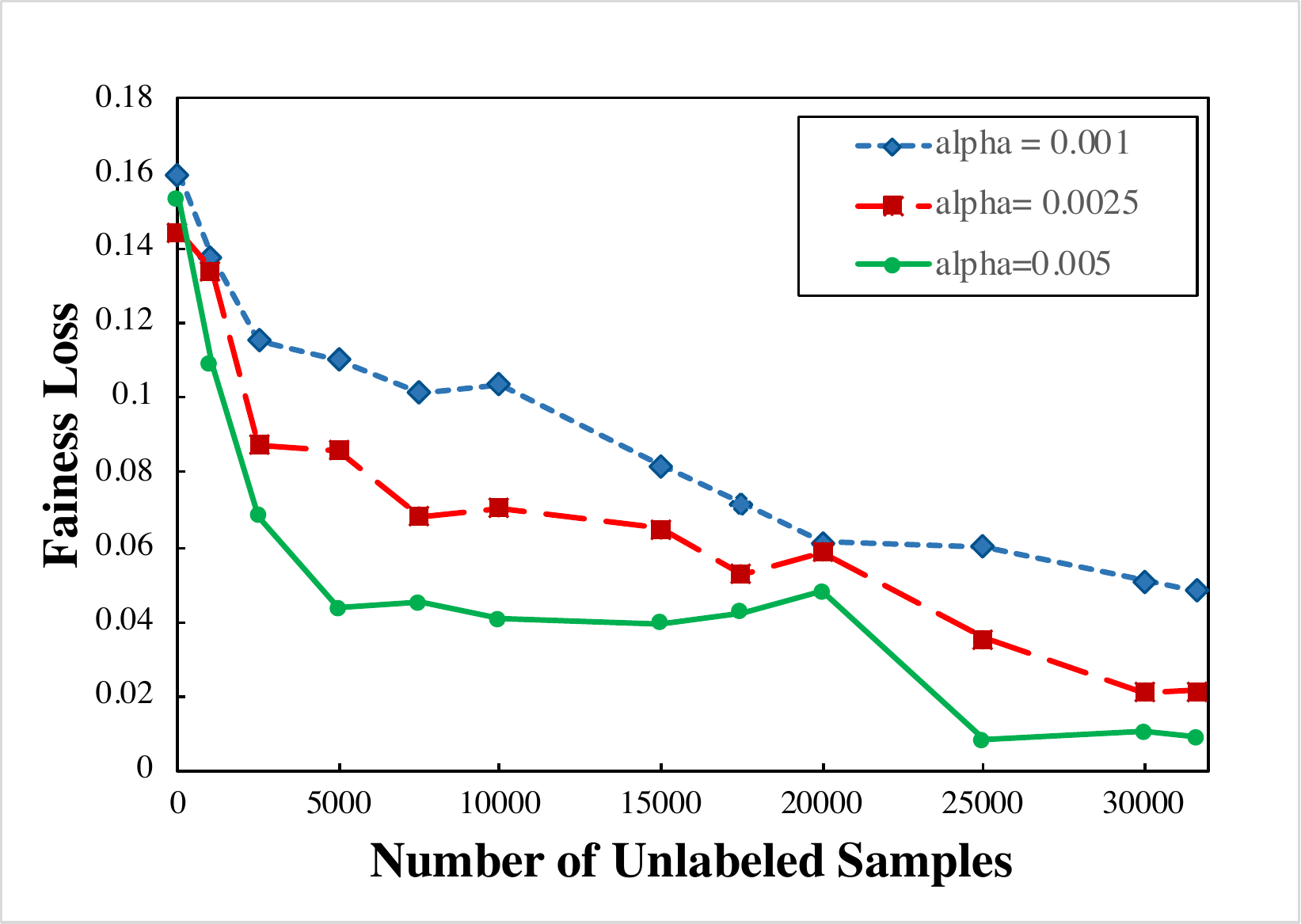}
    \caption{The effect of the number of the unlabeled samples on fairness loss (demographic parity).}
    \label{fig:fairnessunlabled}
\end{figure}

\begin{figure}[ht]
    \centering
    \includegraphics[width=0.33\textwidth]{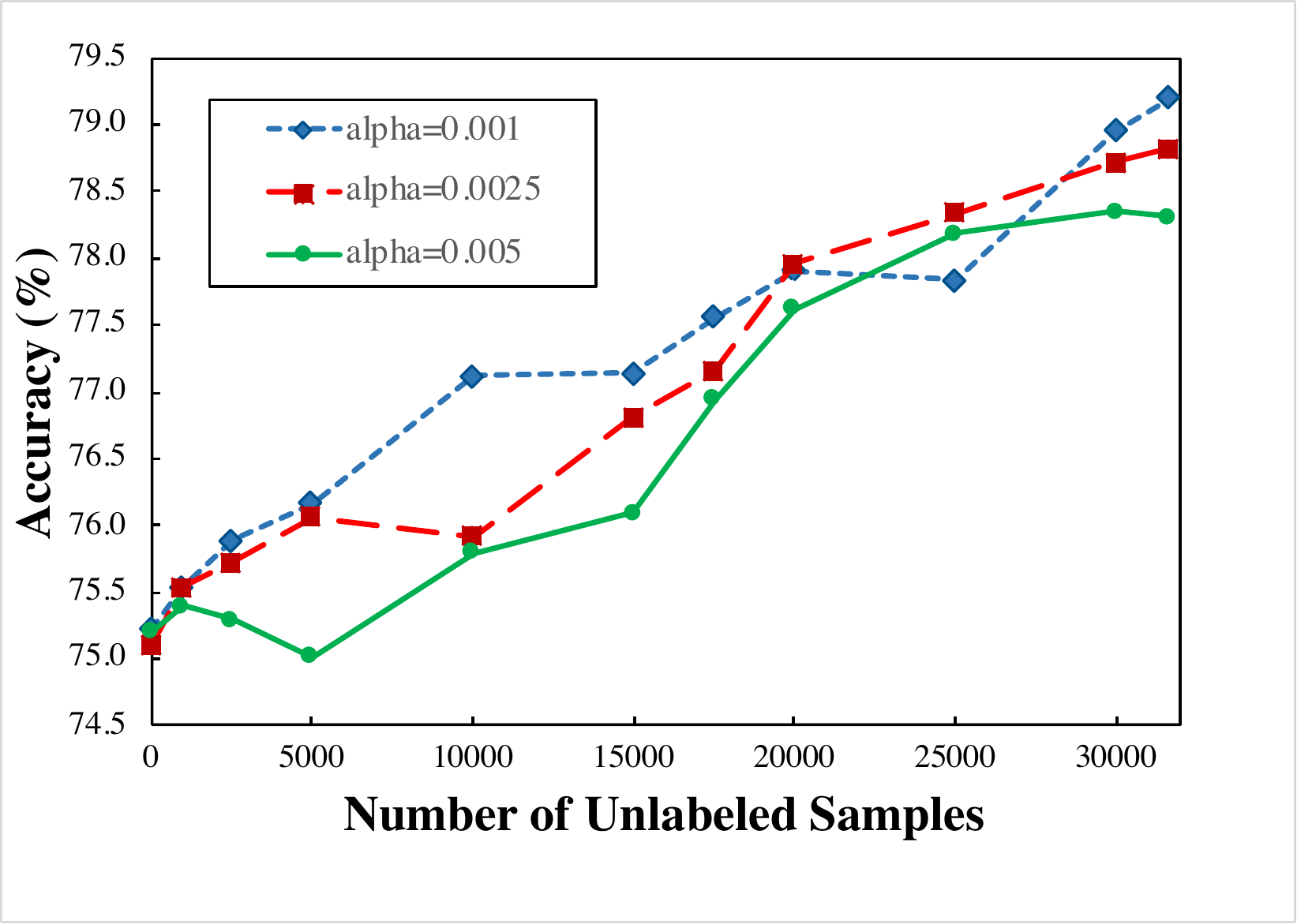}
    \caption{The effect of the number of the unlabeled samples on accuracy.}
    \label{fig:accunlabled}
\end{figure}
\begin{figure*}[ht]
    \centering
    \begin{subfigure}{.33\textwidth}
  \centering
  \includegraphics[width=.88\linewidth]{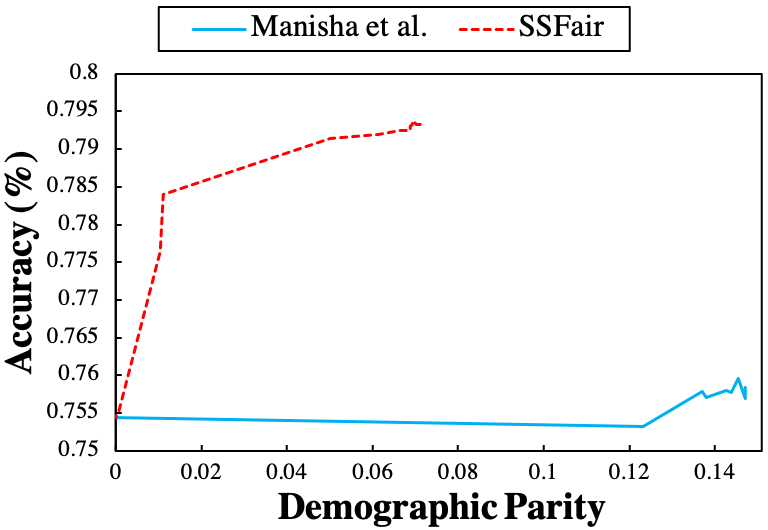}
  \caption{$100$ labeled samples}
  \label{fig1:s100}
\end{subfigure}%
\begin{subfigure}{.33\textwidth}
  \centering
  \includegraphics[width=.88\linewidth]{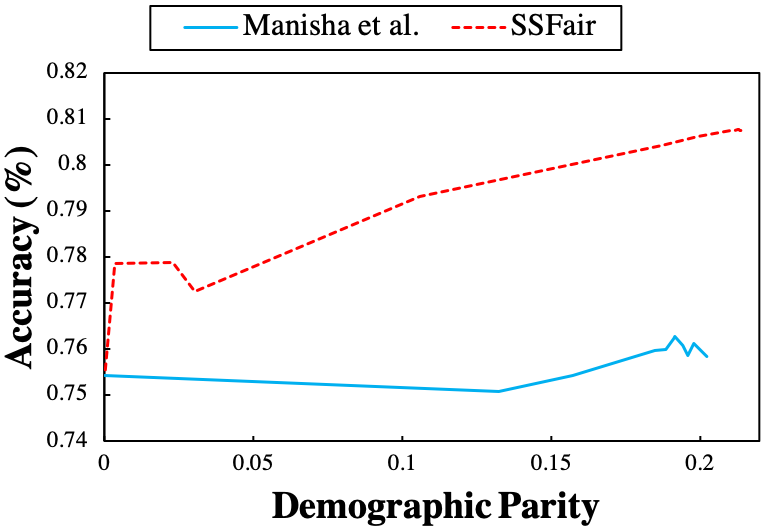}
  \caption{$200$ labeled samples}
  \label{fig1:s200}
\end{subfigure}
%\end{center}
\begin{subfigure}{.33\textwidth}
  %\centering
  \includegraphics[width=.88\linewidth]{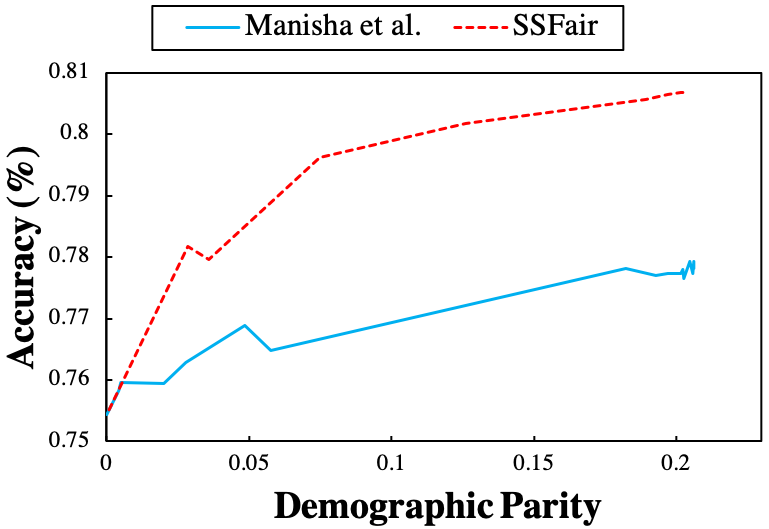}
  \caption{$300$ labeled samples}
  \label{fig1:s300}
\end{subfigure}
    \caption{The trade-off between the demographic parity loss and the accuracy of our proposed model (SSFair) compared to Manisha et al. \cite{manisha2018neural}. The number of labeled samples is $100$, $200$, and $300$ in~\ref{fig1:s100},~\ref{fig1:s200}, and~\ref{fig1:s300}, respectively.}
    \label{fig:dp}
\end{figure*}
\begin{figure*}[ht]
    \centering
    \begin{subfigure}{.33\textwidth}
  \centering
  \includegraphics[width=.88\linewidth]{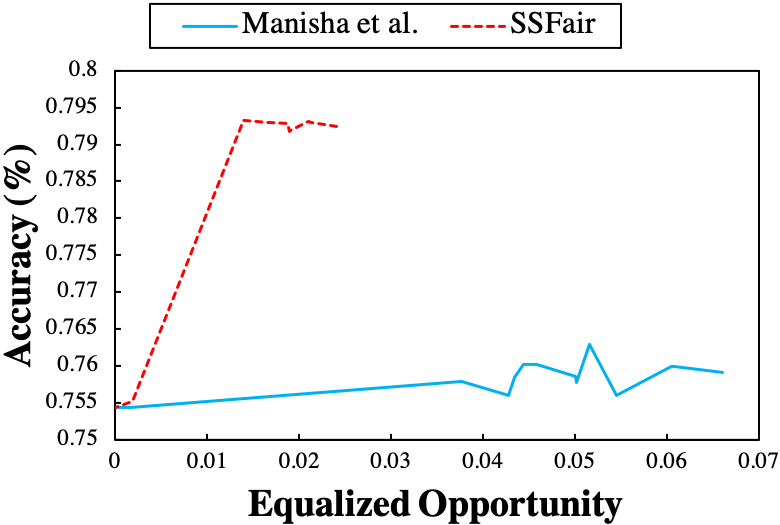}
  \caption{$100$ labeled samples}
  \label{fig2:s100}
\end{subfigure}%
\begin{subfigure}{.33\textwidth}
  \centering
  \includegraphics[width=.88\linewidth]{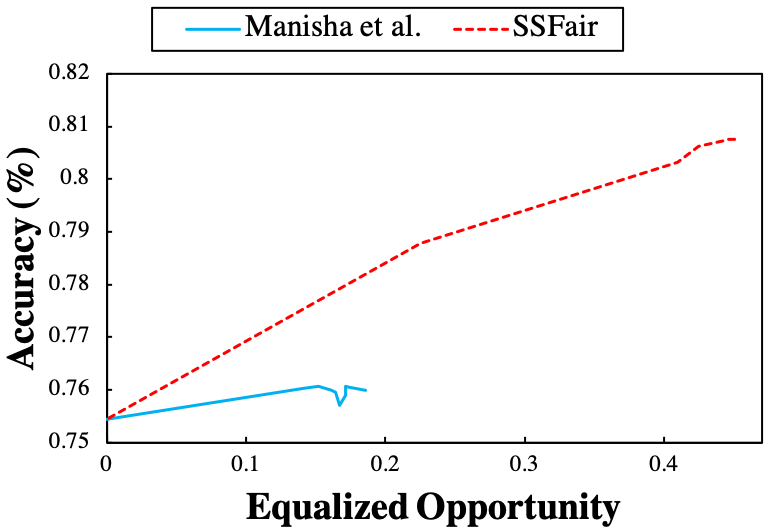}
  \caption{$200$ labeled samples}
  \label{fig2:s200}
\end{subfigure}
%\end{center}
\begin{subfigure}{.33\textwidth}
  %\centering
  \includegraphics[width=.88\linewidth]{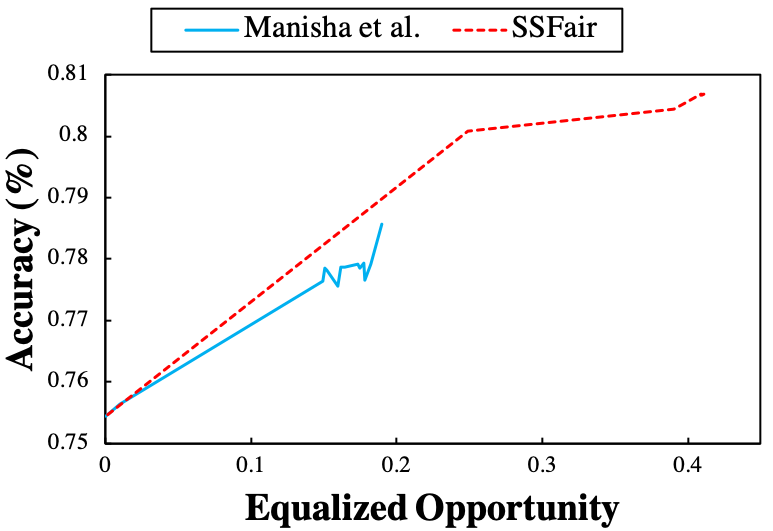}
  \caption{$300$ labeled samples}
  \label{fig2:s300}
\end{subfigure}
    \caption{The trade-off between the equalized opportunity loss and the accuracy of our proposed model (SSFair) compared to Manisha et al. \cite{manisha2018neural}. The number of labeled samples is $100$, $200$, and $300$ in~\ref{fig2:s100},~\ref{fig2:s200}, and~\ref{fig2:s300}, respectively.}
    \label{fig:eopp}
\end{figure*}

\begin{figure*}[ht]
    \centering
    \begin{subfigure}{.33\textwidth}
  \centering
  \includegraphics[width=.88\linewidth]{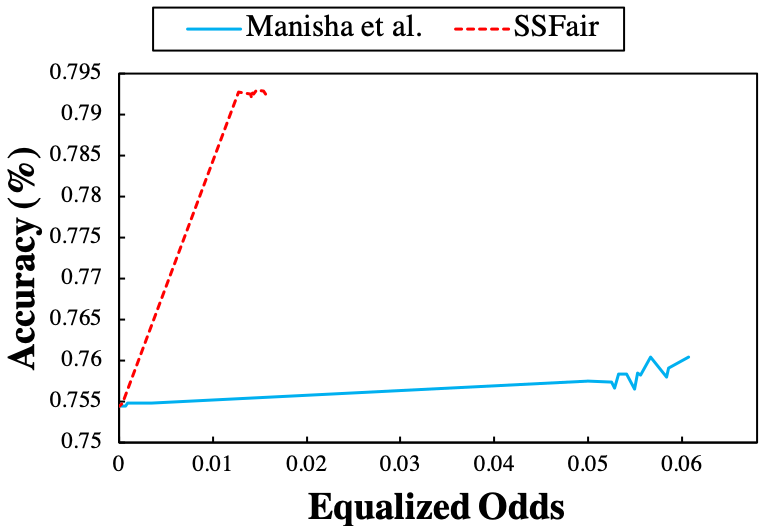}
  \caption{$100$ labeled samples}
  \label{fig3:s100}
\end{subfigure}%
\begin{subfigure}{.33\textwidth}
  \centering
  \includegraphics[width=.88\linewidth]{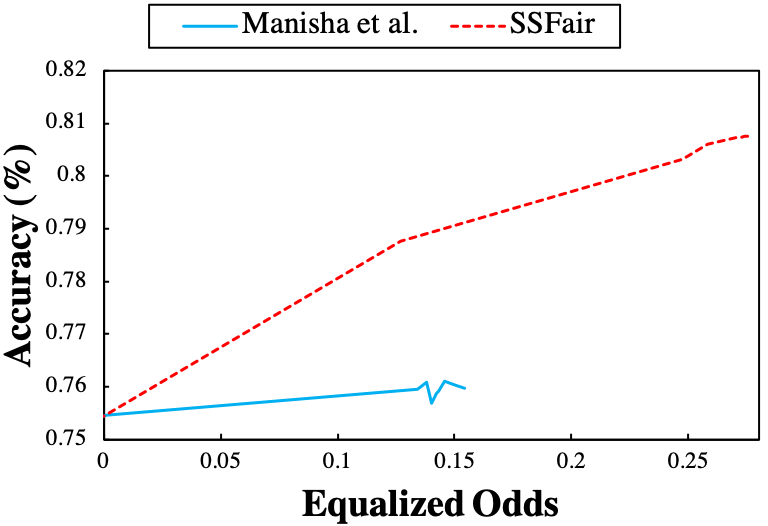}
  \caption{$200$ labeled samples}
  \label{fig3:s200}
\end{subfigure}
%\end{center}
\begin{subfigure}{.33\textwidth}
  %\centering
  \includegraphics[width=.88\linewidth]{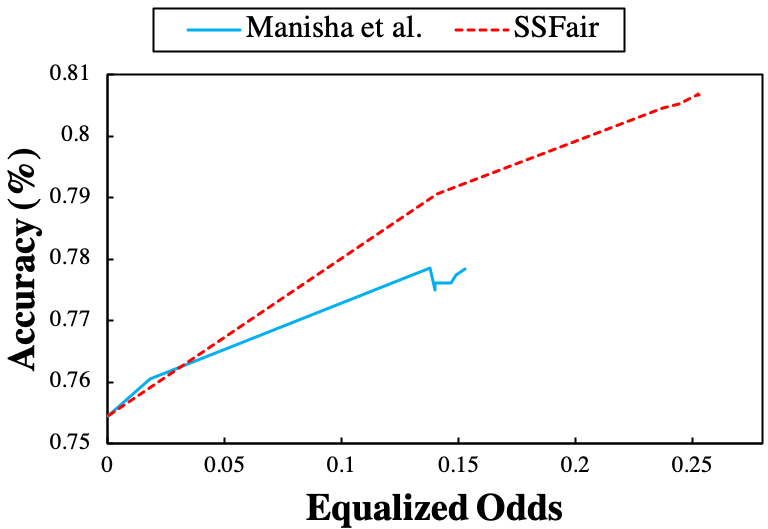}
  \caption{$300$ labeled samples}
  \label{fig3:s300}
\end{subfigure}
    \caption{The trade-off between the equalized odds loss and the accuracy of our proposed model (SSFair) compared to Manisha et al. \cite{manisha2018neural}. The number of labeled samples is $100$, $200$, and $300$ in \ref{fig3:s100}, \ref{fig3:s200}, and \ref{fig3:s300}, respectively.}
    \label{fig:eodd}
\end{figure*}

\section{Experiments}
\label{expsec}
We evaluate and study our proposed model for the fairness problem. We provide experimental results to support our claim that
employing our semi-supervised approach based on neural networks improves
the accuracy and fairness for classification task.
\subsection{Dataset}
We use the UCI Adult Income Dataset (ADULT)\cite{adultdataset,kohavi1996scaling} and study the task of predicting whether a person makes more than $50K$ or not. This is one of the most commonly used benchmarks for evaluating classification approaches for fairness. The proportion of high income individuals across the two groups of men and women are not equal, and therefore there is no demographic parity in the dataset. The dataset has 12 features including categorical and continuous features. 

Categorical features are encoded using one-hot encoding. The age feature is bucketized at the boundaries [18, 25, 30, 35, 40, 45, 50, 55, 60, 65]. The "Sex" feature is considered as a protected feature. We have also filtered out the samples with missing values. The post-processed dataset contains $45,222$ samples with $112$ features. We randomly chose $70\%$ of the samples for the train set and left the rest for the test set.

\subsection{Experimental Setting}
We compare the results of our work with the model proposed by Manisha et al. \cite{manisha2018neural} which is a model based on neural networks to address the fairness problem. To the best of our knowledge, it is the only work done on the trade-off between fairness and accuracy using neural networks. This model \cite{manisha2018neural} is fully supervised and is only trained on the labeled samples.

The hyperparameters of our proposed algorithm are tuned with validation on a randomly selected $20\%$ of the training data. After setting the hyperparameters, the model is trained on the full training set. Eventually, the results on the test data are reported in the experiments.

In our experiments, for both SSFair and Manisha et al. \cite{manisha2018neural}, a Multilayer Perceptron (MLP) neural network with $1$ hidden layer of size $32$ is used to model the function $f(X)$. Rectified Linear Unit (ReLU) activation is used for the outputs of the hidden layer. Since the task is binary classification, we use sigmoid function as the activation function on the last layer and get the final output as the result of that. A dropout layer with a dropout rate of $20\%$ is used after the hidden layer. The regularization parameter $\beta$ is selected from $\{10^{-5}, 10^{-4}, 10^{-3}, 10^{-2}, 10^{-1}, 1.0\}$ for each experiment based on the results of the validation process. Finally, the confident degree parameter $\lambda$ is set to $0.99$ for SSFair.

We use the Adam optimizer \cite{kingma2014adam} to train the models. We choose the learning rate of $10^{-3}$ and use the default values recommended in \cite{kingma2014adam} for the other parameters of the optimizer. Training the neural networks is done by running Adam over $1000$ epochs of training data, when using shuffled mini-batches of size $512$.
\subsection{Experimental Results}
In this section, we present the results of our experiments on the ADULT dataset to demonstrate the effectiveness of our semi-supervised learning approach for the fairness problem.
\vspace{0.5mm}
\subsubsection{\textbf{The effect of unlabeled data on accuracy and fairness}}
We would like to verify that using unlabeled data can help our algorithm to improve on both aspects of accuracy and fairness. We performed experiments by increasing the number of unlabeled samples while keeping the number of labeled samples fixed to $100$ to investigate the effect of adding unlabelled data. We have experimented with three different values of $\{0.001, 0.0025, 0.005\}$ for parameter $\alpha$.

The plot of fairness loss versus the number of unlabeled samples is illustrated in Figure \ref{fig:fairnessunlabled}. For calculating the fairness loss, the output of the classifier ($\hat{y}$) is binarized with the threshold of $0.5$ to provide a binary outcome. Demographic parity is selected as the fairness loss in this part. As these plots suggest, fairness loss improves as we increase the size of the unlabeled set (note that higher fairness is achieved with fairness loss is lower). This experiment verifies that fairness in our model can benefit from unlabeled data and therefore our approach has been successful in utilizing unlabeled data to improve fairness.
Moreover, we paid special attention to the existing trade-off between accuracy and fairness as well. Particularly, we were interested in understanding whether the improvement in fairness by increasing the size of unlabeled data could be a result of potential losses on the accuracy. To understand this effect, the plot of  accuracy versus the number of unlabeled samples is also illustrated in Figure \ref{fig:accunlabled}. As it is clear from the plots, the accuracy of the classifier increases as we grow the number of unlabeled samples as well. This result validates that our approach provides a solution for using additional unlabeled data to improve both factors of accuracy and fairness.
\vspace{1mm}
\subsubsection{\textbf{Comparison against fully supervised approach}}
We demonstrate the benefit of our semi-supervised learning approach for the fairness problem versus a full supervised model. We experimented with a varying number of labeled samples ($100$, $200$, and $300$). For experiment with $n$ labeled samples, we randomly chose $n$ samples from the training set and kept their ground truth label while we changed the label of the other samples to $u$.

The results are illustrated in Figures \ref{fig:dp}, \ref{fig:eopp}, and \ref{fig:eodd}. Different points on the curves are obtained by using different values for parameter $\alpha$ which is varied from $10^{-7}$ to $10^{4}$ to impose different levels of fairness on the classifier. Generally, there exists a trade-off between accuracy and fairness and parameter $\alpha$ controls this trade-off: increasing $\alpha$ would result in increasing the accuracy while decreasing the fairness. For each value of $\alpha$, the experiment is repeated five times and the averaged results are reported.

Comparing two algorithms, one which can produce higher accuracy while maintaining the same level of fairness loss is considered the superior one. As it is evident from the results, SSFair provides higher accuracy for the same level of fairness loss compared to the approach of Manisha et al. This conclusion is consistent for all of the three fairness measurements, suggesting the effectiveness of exploiting unlabeled data by using semi-supervised learning for the fairness problem. It is worth noting that the effect of using unlabeled data is more evident in cases with fewer labeled samples, which indicates that this approach is most helpful in scenarios with scarce labeled data.

Our understanding of such behavior is that since unlabeled data does not include any label information, they do not hold biased information for the labels either. Therefore, they can be beneficial not only to the accuracy but also to the fairness of the classifier. Our experiments show that SSFair is capable of exploiting the structure and information of unlabeled data to increase the accuracy and fairness compared to a fully supervised model.

\section*{Conclusion}
We proposed a classifier based on neural networks for semi-supervised learning to tackle the fairness problem. The proposed model, SSFair, employs the Pseudo-labeling approach to exploit the information in the unlabeled data. We studied the effect of using unlabeled data on learning a fair classifier, and showed with our experiments that unlabeled data could be beneficial not just for the accuracy but also for the fairness. SSFair is evaluated on three fairness measurements, demographic disparity, equalized opportunity, and equalized odds. In the experiments, it is demonstrated that semi-supervised learning can achieve higher fairness and accuracy compared to the approaches that only use the labeled data.

\bibliographystyle{IEEEtran}
\bibliography{SSFair}
\end{document}